*Research Article*

# OpenSync: An open-source platform for synchronizing multiple measures in neuroscience experiments

Moein Razavi [1,2], Vahid Janfaza [2], Takashi Yamauchi [1 *], Anton Leontyev [1], Shanle Longmire-Monford [1], Joseph Orr [1]

[1] Department of Psychological and Brain Sciences, Texas A&M University
[2] Department of Computer Science and Engineering, Texas A&M University
* Correspondence: moeinrazavi@tamu.edu

**Abstract:**
*Background:* The human mind is multimodal. Most behavioral studies rely on century-old measures such as task accuracy and latency. To better understand human behavior and brain functionality, we need to analyze physiological and behavioral signals of various sources. However, it is technically complex and costly to design and implement experiments that record multiple measures. To address this issue, a platform that synchronizes multiple measures is needed.

*Method:* This paper introduces an open-source platform named OpenSync, which can be used to synchronize numerous measures in neuroscience experiments. This platform helps to automatically integrate, synchronize and record physiological and behavioral signals (e.g., electroencephalogram (EEG), galvanic skin response (GSR), eye-tracking, body motion, etc.), user input response (e.g., from mouse, keyboard, joystick, etc.), and task-related information (stimulus markers). In this paper, we explain the features of OpenSync and provide two case studies in PsychoPy and Unity.

*Comparison with existing tools:* Unlike proprietary systems (e.g., iMotions), OpenSync is free and easy to implement, and it can be used inside any open-source experiment design software (e.g., PsychoPy, OpenSesame, Unity, etc., *https://pypi.org/project/OpenSync/* and *https://github.com/TAMUCogLab/OpenSync*).

*Results:* Our experimental results show that the OpenSync platform is able to synchronize multiple measures with microsecond resolution.

**Keywords:** multiple measures synchronization; automatic device integration; open-source; PsychoPy; Unity

## 1. Introduction

*1.1. Problem Statement*

The brain supervises different autonomic functions such as cardiac activity, respiration, perspiration, etc. Current methods to study human behavior include self-report, observation, task performance, gaze, gait and body motion, and physiological measures such as electroencephalogram (EEG), electrocardiogram (ECG), and functional magnetic resonance imaging (fMRI). Human behavior is inherently multimodal and interdependent; however, previous studies in main used a single measure and did not investigate the connections between various measures [1]. Multimodal experiments are important because measures from various sources have a location (spatial) or timing (temporal) overlap in the brain [2]. Hochenberger suggests that the experiments that employ multiple measures enhance our understanding of senses that operate in parallel [2]. Multiple measures are more informative than a single measurement because it distinguishes between different functions in the mind and allows for a more accurate interpretation of behavior [4]. In other words, in a similar way that using numerous features improves accuracy in classification tasks, combining different measures helps to improve the predictions of human behavior in experiments that study brain functionality and its association with behavior [5].

*1.2. Studies with multimodal measures*

Recently, several studies have been conducted using the integration of multiple measures into the experimental psychology/neuroscience portal. Reeves et al. emphasizes the importance of using multiple measures in Augmented Cognition (AUGCOG) [6]; they argue that combining multiple measures improves Cognitive State Assessment (CSA). Jimenez-Molina et al. shows that analyzing electrodermal activity (EDA), photoplethysmogram (PPG), EEG, temperature, and pupil dilation at the same time significantly improves the classification accuracy in a web-browsing workload classification task, compared to using a single measure [7].

Born et al. uses EEG, GSR, and eye-tracking to predict task performance in a task load experiment; they find that low-beta frequency bands, pupil dilations, and phasic components of GSR are correlated with task difficulty [8]. They also show analyzing EEG and GSR together are statistically more reliable than individual analyses. Wang et al. uses PsychoPy, EEG, and Lab Streaming Layer (LSL) for Brain-Computer Interface (BCI) stimulus presentations. They use these to synchronize stimulus markers and EEG measurements [9]. Leontyev et al. combines user response time and mouse movement features with machine learning techniques and finds improvement in the accuracy of predicting attention-deficit/hyperactivity disorder (ADHD) [10]–[12]. Yamauchi et al. combines behavioral measures and multiple mouse/cursor motion features to predict people's emotions and cognitive conflict in computer tasks [13], [14]. Yamauchi et al. further demonstrates that people's emotional experiences change as their tactile senses (touching a plant) are augmented with visual senses ("seeing" their touch) in a multisensory interface system [15]. Razavi et al. provides a comprehensive tutorial on how to use LSL for multimodal experiments using PsychoPy, and Unity [16]. Chen et al. tried to identify possible correlations between increasing levels of cognitive demand and modalities. They tested this using speech, digital pen, and freehand gesture with eye activity, galvanic skin response, and EEG [17]. Lazzeri et al. employ physiological signals, eye gaze, video, and audio acquisition to perform an integrated affective and behavioral analysis on Human-Robot Interaction (HRI) [18]; they acquire synchronized data from multiple sources and investigate how autistic children can interact with affective robots. Cornelio et al. review current body of knowledge in multisensory integration system and state despite that such systems increase the level of human control on the environment, they bring some technological issues and challenges such as increasing human responsibility in working with such systems [19]. Charles and Nixon review 58 articles on mental workload tasks. They find that physiological measurements such as ECG, respiration, GSR, blood pressure, EOG, and EEG need to be triangulated because though they are sensitive to mental workload, no single measure satisfies in predicting mental workload [20]. Lohani et al. suggest that analyzing multiple measures such as head movement with physiological measures (e.g., EEG, heart rate, etc.) can be used when detecting driver's physiological states (e.g., distraction) [21]. Levin et al. propose the use of EEG, eye-tracking, and image markers in a BCI system to classify images that were difficult for machine learning to classify [22]. They capture various finger interactions when navigating web pages with the vertical surface, including tap and swipe up. Gibson et al. integrate questionnaires, qualitative methods, and physiological measures including ECG, respiration, electrodermal activity (EDA), and skin temperature to study activity settings in disabled youth [23]; they state that using multiple measures reflects a better real-world setting of the youth experiences. Sciarini and Nicholson adopt EEG, eye blink, respiration, cardiovascular activity, and speech measures in a workload task performance [24].

Despite the need for multimodal experiments, systems that are capable of integrating separate measurements are rare. There are several proprietary and non-proprietary platforms (e.g., iMotions [25], BioPac [26], neurobs [27], etc.) that ease the implementation of multimodal experiments; however, these systems have several drawbacks. First, it is costly and challenging to integrate different devices. Second, separate software packages are needed for stimulus presentation and data acquisition because package options are limited in these platforms, making it difficult to synchronize multiple data streams. This may result in undesirable delays as different software may have different processing times. Third, these systems often require the devices to be connected and started one by one, which is time-consuming. An instant and easy experiment setup is essential for collecting data from a large group of participants, making these systems impractical. Finally, the lack of proper organization in the data files makes it difficult to store and access data from multiple measurements in a format suitable for data analysis.

In this paper, we present OpenSync, a system that allows stimulus presentation, data acquisition, and recording all in the same software, and demonstrate how to make this process automatic and adaptable for various experiments. The following section compares OpenSync to existing competitor software for creating experiments with multiple measures.

*1.3. Comparison to Existing Tools*

In this section, we first introduce available platforms for data acquisition in psychology and neuroscience experiments and compare them to OpenSync (Table 1).

- iMotions is a proprietary platform used to implement human behavior experiments by integrating multiple biometric sensors (EEG, eye-tracking, EDA, etc.) [25]. iMotions allows for the integration of the devices that are available for the platform. It also uses a built-in feature that helps stimulus presentation. However, iMotion's stimulus presentation capabilities are limited—not many custom and open-source devices can be integrated with iMotions. The system does not have an automatic recording function.

- BioPac is another proprietary system that provides multiple recordings of EEG, eye-tracking, GSR, etc., along with data acquisition and analysis tools used for behavioral experiments [26]. BioPac needs external hardware for synchronization and stimulus presentation. BioPac also has a limited list of devices that are compatible with the platform and has no automatic recording function.

- Naturalistic Experimental Design Environment (NEDE) is a platform for Unity engine that integrates eye-tracking and EEG devices for experiments in the 3D environment [28]. NEDE does not have automatic recording capabilities, making the implementation of a large-scale experiment difficult.

- The Unified Suite for Experiments (USE) includes a set of hardware and software tools to integrate EEG and eye-tracking for behavioral neuroscience experiments [29]. Though it is comprehensive, the system lacks automatic recording capabilities.

NEDE and USE both need external hardware and software for synchronization and stimulus presentation. Moreover, they work only with specific devices, and it is not straightforward to add new devices to these platforms.

- Neurobehavioral Systems Presentation® (Neurobs Presentation) is a neuroscience stimulus delivery and experiment control tool for auditory, visual, and multimodal stimulus presentation and data acquisition. The system is capable of handling behavioral, psychological, and physiological experiments, including fMRI, EEG, eye movements, single-neuron recording, and response time measurements [27]. Neurobs Presentation uses LSL to synchronize available devices for the platform but does not allow the user to add other devices. For stimulus presentations, Neurobs Presentation permits custom programming with Printer Command Language (PCL) and Python language; however, stimulus presentation features are limited, especially with the Python platform.

Table 1 provides a comparison between OpenSync and other platforms. OpenSync outperforms proprietary systems such as iMotion and BioPac in its extendibility; OpenSync also surpasses free/open-source systems, NEDE and USE, in its ease of implementation and flexibility. For example, NEDE and USE support JavaScript and C# for extension while OpenSync allows Python, C, Java, C#, C++, and Matlab. Proprietary systems, iMotions and BioPac, are relatively easy to implement, but they do not offer programming capacity. OpenSync also offers easy user extendibility in that the manual specifies how to extent the platform. It is also the only one with extensive device support with thorough documentations.

Table 1. Platform Comparisons

| | iMotions | BioPac | NEDE | USE | Neurobs Presentation | **OpenSync** |
|---|---|---|---|---|---|---|
| Free/Open-source | ✗ | ✗ | ✓ | ✓ | ✗ | ✓ |
| Multi-Device Synchronization | ✓ | ✓ | ✓ | ✓ | ✓ | ✓ |
| Supported Programming Languages | ✗ | ✗ | JavaScript | C# | PCL, limited Python programming | Python, C, C#, C++, Java, Matlab |
| Device Support | Limited[1] | Limited | Limited | Limited | Limited | Broad[2] |
| Comprehensive | ✓ | ✗ | ✗ | ✗ | ✓ | ✓ |
| Easy Implementation | ✓ | ✓ | ✗ | ✗ | ✓ | ✓ |
| No Extra Hardware | ✓ | ✗ | ✗ | ✗ | ✓ | ✓ |
| Automatic Record | ✗ | ✗ | ✗ | ✗ | ✗ | ✓ |
| User Extendable | ✗ | ✗ | Difficult[3] | Difficult | ✗ | Easy[4] |
| Portable | ✗ | ✗ | ✗ | ✗ | ✗ | ✓ |
| Used for VR Experiments | ✗ | ✗ | ✓ | ✓ | ✗ | ✓ |
| Latency Resolution | N/A | 1 ms [30] | 0.32 ms [28] | 0.1 ms [29] | 0.1 ms [31] | < 0.07 ms |

Note: [1]Limited: the systems that accept the devices made by licensed manufacturers only are classified as "Limited". For example, eye-tracking data can be integrated into BioPac but BioPac allows only licensed manufacturers such as Argus Science, EyeTech, and Arrington Research, but not Tobii and Gazepoint. [2]Broad: the system that allows many third-party custom-made devices (e.g., the devices implemented in Arduino and Arduino shields and/or the OpenBCI framework as well as off-the-shelf consumer grade devices—Muse and Mindwave for EEG) is classified as "Broad." [3]Difficult: we classified the system as "Difficult" if the system does not have proper documentation for device extension. [4]Easy: we classified the system as "Easy" when the system has a guide on extension in its manual, Github, wiki, and/or sample scripts/demos (e.g., https://github.com/TAMUCogLab/OpenSync/wiki/Extending-OpenSync-for-New-Devices)

OpenSync has good latency resolution as well. The last row of Table 1 shows the latency resolution reported by different platforms for data synchronization. Latency resolution denotes the discrepancy between the time that data is sent from the device and the time the same data is saved on the disk. For BioPac and Neurobs Presentation, this information was obtained from their website. For NEDE and USE, this information is obtained from the articles introducing these platforms [28, 29]. OpenSync outperforms the latency resolution reported by the other platforms for synchronization (0.1 ms for Neurobs Presentation) by more than 30%. A detailed explanation about the experiment for measuring OpenSync latency resolution is available in Section 4 of this paper. In terms of synchronization, one important feature that distinguishes OpenSync from other platforms is that OpenSync runs and records data in the background of the stimulus presentation software and its latency of the stimulus presentation software does not affect the performance and latency of OpenSync.

This paper is structured as follows. Section 2 introduces the method we used in developing the OpenSync platform. Section 3 describes two case studies of using OpenSync in PsychoPy and Unity. Section 4 tests the applicability of OpenSync via a time synchronization test. The final section of the paper provides the discussion and possible future extensions for OpenSync. Our GitHub, *https://github.com/TAMUCogLab/OpenSync*, provides an overview, installation procedures, demo, data analysis procedures, API reference and more about OpenSync (*https://pypi.org/project/OpenSync/*).

## 2. Method

*2.1. Development Criteria*

Below, we present the criteria we considered in developing OpenSync to make it useful for human behavior research experiments.

- **Multi-Device Synchronization** – synchronize several sensors, I/O devices, and markers with *high temporal* resolution
- **Easy implementation** – the package provides a set of libraries and customized device Software Development Toolkits (SDKs) imported into popular psychology/neuroscience experiment design software packages (e.g., PsychoPy). Only one computer is needed for connecting devices, running experiments, and recording data.
- **Quick implementation** – integration of multiple measures from multiple sources can be done in a short time without any programming knowledge.
- **Practical** – this software package addresses key challenges in synchronizing multiple sensors for psychology/neuroscience experiments, such as being time-consuming to set up, synchronize, and record data from numerous resources. It can also synchronize data from multiple computers using a Local Area Network (LAN) connection.
- **Modular** – the data from all the devices and markers can be recorded independently using the individual functions in the proposed platform.
- **Comprehensive** – data acquisition and stimulus presentation can be applied in the same software.
- **Portable** – can be used within open-source psychological experiment design software (e.g., PsychoPy, OpenSesame, Unity).
- **Automatic** – recording data from multiple sources and saving them on the disk will be done automatically with minimal manual intervention.
- **Adjustable** – the platform is expandable to support new devices and update the current devices' SDKs. Also, all open-source devices, as well as non-open-source devices that can stream data with one of the open-source platforms (e.g., C, C++, C#, Python, Java, and Octave) can be synchronized.
- **Open source and Cost-efficient** – it is a free open-source software package and does not need any extra intermediate hardware.
- **Multiplatform** – the software package is not dependent on the operating system since it has been developed by open-source platforms (i.e., Python, C++, and C#)
- **Offline simulation** – prerecorded data can be loaded instead of the data from actual devices. For this paper, we focused on synchronization. Offline simulation is not in the scope of this manuscript, and we leave it for a future study.

*2.2. OpenSync platform*

In this section, we provide the details of the OpenSync platform. Figure 1 illustrates the layout of the OpenSync platform. OpenSync employs a modular design; it consists of two main modules—Synchronizer and Recorder—and four submodules—I/O devices, Sensors, Controller, Marker. The Controller is responsible for initializing and synchronizing data streams and sending data to the Recorder module. It uses Lab Streaming Layer (LSL) as its core protocol for data streaming and synchronization [32]. The I/O module processes user response data from I/O devices (Keyboard, Mouse, Joystick, etc.). The Markers module reads extraneous information such as stimulus ID, experimental conditions, subject profiles) from user defined input files and/or programmatically generated manipulation protocols (e.g., adaptive stimulus presentation duration). Finally, the Sensors module is used for streaming data from various biological sensors (EEG, GSR, eye-tracking, body motion, etc.). The Recorder module records all the streams from different submodules in a single file with Extensible Data Format (*.xdf* extension). Table 2 describes the pseudocode for the

overall structure of OpenSync. For an extended demo, please see
*https://github.com/TAMUCogLab/OpenSync/wiki/Quick-Start-(Demo)*.

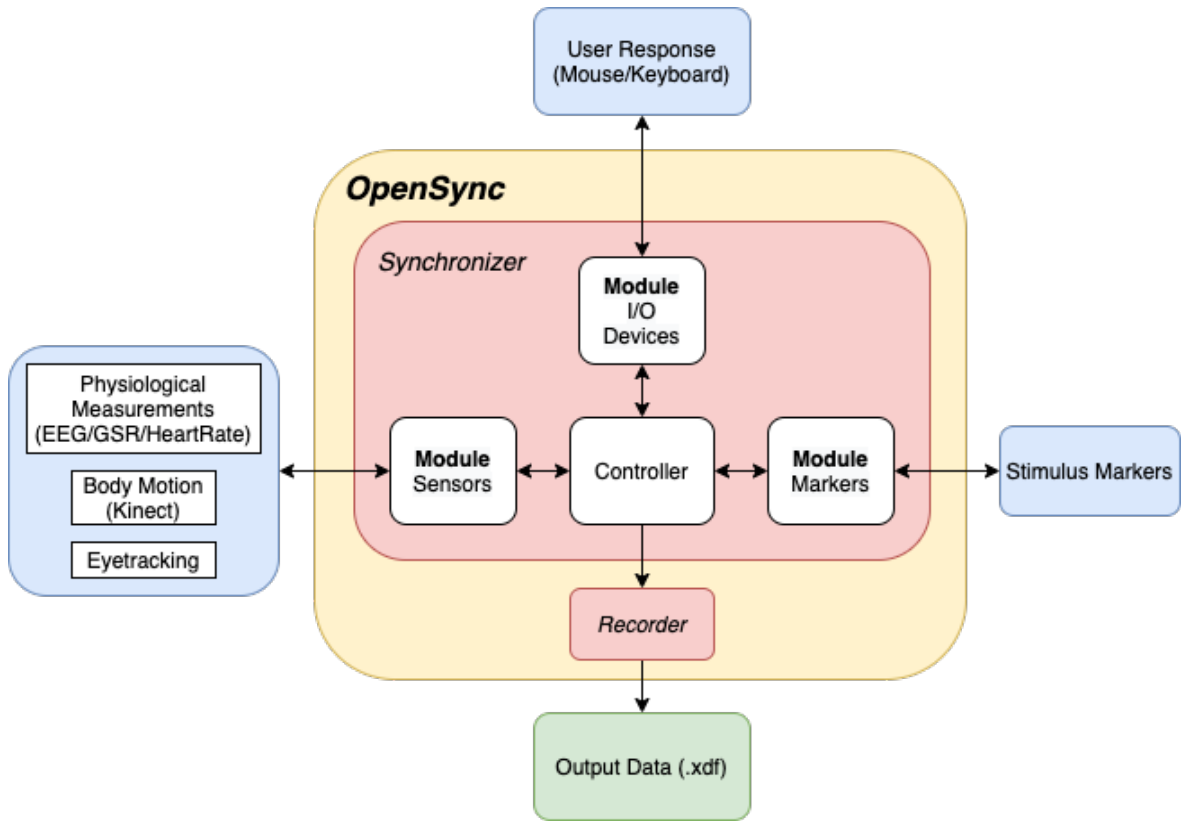

Figure 1. Overall view of OpenSync platform. Different modules and the dataflow between them are depicted in this figure.

Table 2. Pseudocode for the overall structure of OpenSync

| | **OpenSync Platform** |
|---|---|
| 1 | # **Include** required libraries (e.g., LSL) |
| 2 | # **Include** required device SDKs |
| 3 | **Module** OpenSync(): |
| 4 | _Config_= Get User Input Parameters<br>// Synchronizer module (Lines 5-7) |
| 5 | Call class I/O device from module I/O devices (_Config_)   // Call for each I/O device |
| 6 | Call class Stimulus Markers from module markers (_Config_) |
| 7 | Call class Physiological Sensors from module sensors (_Config_)   // Call for each physiological sensor |
| 8 | Initialize XDF Recorder (_Config_)   // Recorder module |
| 9 | Start Synchronization and Recording Data |
| 10 | **Module** I/O Devices: |
| 11 | **Class** I/O device:   // Contain class Keyboard, class Mouse, class Joystick, class Gamepad |
| 12 | **Def** __init__(_Config_): |
| 13 | Create and initialize I/O device data stream |
| 14 | **Def** stream_data(I/O object): |
| 15 | Stream I/O data |
| 16 | **Module** Markers: |

| | | |
|---|---|---|
| 17 | **Class** Stimulus Markers: | |
| 18 | **Def** __init__(_Config_): | |
| 19 | Create and initialize stimulus marker data stream | |
| 20 | **Def** stream_marker(marker): | |
| 21 | Stream stimulus marker (int/string) | |
| 22 | **Module** Sensors:   // Contains libraries of different physiological sensors | |
| 23 | **Class** Physiological Sensors:   // Contains class EEG, class BodyMotion, class EyeTracking | |
| 24 | **Def** __init__(_Config_): | |
| 25 | Initialize physiological sensor and stream data | |

Table 3 describes OpenSync's built-in functions. Functions 2-6 perform the initialization, configuration, and streaming of EEG devices. For the OpenBCI Cython device, if the Daisy module is installed, we can set daisy=True to record Daisy's data. We should specify the USB port to which the OpenBCI Bluetooth dongle is connected (default = "COM3"). In BrainProducts LiveAmp device, the number of EEG channels (n_channels) can be set on {16, 32, 64} and the sampling frequency (sfreq) can be set on {250, 500, 1000}.

Function 10 in Table 3 runs the Gazepoint Control program and streams Gaze data. If the Gazepoint biometric device is connected, the biometrics data can be streamed by setting biometrics=True. In function 12, by setting clickable_object, pos, and click parameters, we can initialize streams for the name of the clicked object, the mouse position coordinates, and the name of the mouse button clicked (left, middle, or right), respectively.

In function 13, by setting keypress and pos parameters, we can initialize streams for the name of the pressed key and the joystick position coordinates, respectively.

Detailed information about OpenSync functions and demo are available at: *https://github.com/TAMUCogLab/OpenSync*

Table 3. Built-in OpenSync functions (the functions in bold text are used for initialization of devices; other functions are used for streaming data from different devices)

| | **Functions for Physiological Sensors** | **Description** |
|---|---|---|
| 1 | **EEG = OpenSync.sensors.EEG()** | Initialize, configure and stream EEG devices data |
| 2 | EEG.OpenBCI_Cyton(port="COM3", daisy=False) | |
| 3 | EEG.Unicorn() | |
| 4 | EEG.LiveAmp(n_channels, sfreq) | |
| 5 | EEG.Mindwave() | |
| 6 | EEG.Muse() | |
| 7 | **Body_Motion = OpenSync.sensors.BodyMotion()** | Initialize and stream Kinect body motion data |
| 8 | Body_Motion.KinectBodyBasics() | |
| 9 | **Eye_Tracking = OpenSync.sensors.EyeTracking()** | Initialize, configure and stream Gazepoint data |
| 10 | Eye_Tracking.Gazepoint(biometrics=False) | |
| 11 | **GSR = OpenSync.sensors.GSR()** | Initialize, configure and stream GSR data |
| 12 | GSR.eHealth(port="COM3") | |
| | **Functions for I/O Devices** | |
| 11 | **keyboard = OpenSync.i_o.Keyboard(Name)** | Create keyboard object for streaming data |
| - | keyboard.stream_keypress (PsychoPy_keyboard_object) | Stream the name of the pressed key |
| 12 | **mouse = OpenSync.i_o.Mouse(Name, clickable_object=True, position=True, click_type=True)** | Create mouse object for streaming data |
| - | mouse.stream_clicktype(PsychoPy_mouse_object) | Stream the type of click {left, middle, right} |
| | - mouse.stream_click(PsychoPy_mouse_object) | Stream the name of the clicked object |
| | - mouse.stream_pos(PsychoPy_mouse_object) | Stream the position of the mouse (in every frame) |
| 13 | **joystick = OpenSync.i_o.Joystick(Name, keypress=True, pos=True)** | Create joystick object for streaming data |
| | - joystick.stream_keypress(PsychoPy_joystick_object) | Stream the keypress event |
| | - joystick.stream_pos(PsychoPy_joystick_object) | Stream the position of the joystick |
| 14 | **gamepad = OpenSync.i_o.Gamepad(Name)** | Create gamepad object for streaming data |
| | - gamepad.stream_buttonpress(PsychoPy_gamepad_object) | |
| | **Function for Stimulus Markers** | |
| 15 | **marker_obj = OpenSync.markers.marker(Name)** | Create marker object for streaming markers |
| | - marker_obj.stream_marker(_marker) | Stream the stimulus marker (int/string) |
| | **Function for Recording Data on Disk** | |
| 16 | OpenSync.record_data("File Address") | Record all streams in a file with *.xdf* extension |

Table 4 shows the list of libraries and device APIs for different biosensors that we customized for OpenSync. As Table 4 shows, OpenSync allows off-the-shelf consumer-grade devices (e.g., Neurosky, Muse, Kinect), as well as research-grade high-capacity devices (BrainProducts LiveAmp, OpenBCI Cyton).

Table 4. Customized libraries and device APIs used in OpenSync

| Type | Device | Source Language | Library and Functions |
|---|---|---|---|
| EEG | g.tec Unicorn | C++ | https://github.com/TAMUCogLab/OpenSync/tree/master/Source Files/Unicorn |
| | BrainProducts LiveAmp | C++ | https://github.com/TAMUCogLab/OpenSync/tree/master/Source Files/LiveAmp |
| | OpenBCI Cyton (+Daisy) | Python | https://github.com/TAMUCogLab/OpenSync/tree/master/Source Files/OpenBCI |
| | NeuroSky Mindwave | --- | https://pypi.org/project/mindwavelsl/ |
| | Muse | --- | Link 1: BlueMuse_Application<br>Link 2: BlueMuse_Installation_Guide |
| GSR | eHealth Sensor v2.0 Arduino Shield | C, Python | https://github.com/TAMUCogLab/OpenSync/tree/master/Source Files/eHealth |
| | Gazepoint Biometrics Device | Python | https://github.com/TAMUCogLab/OpenSync/tree/master/Source Files/Gazepoint |
| Eye-tracking | Gazepoint | Python | https://github.com/TAMUCogLab/OpenSync/tree/master/Source Files/Gazepoint |
| | Tobii for HTC VIVE | C++ | https://github.com/TAMUCogLab/OpenSync/tree/master/Source Files/TobiiPro_SRanipal |
| Body Motion | Kinect | C++ | https://github.com/TAMUCogLab/OpenSync/tree/master/Source Files/Kinect-BodyBasics |

## 3. Case Studies

*3.1. Study 1: Using OpenSync in PsychoPy*

Figure 2 shows a code snippet of OpenSync implemented in PsychoPy. For details and demo, please visit *https://github.com/TAMUCogLab/OpenSync/wiki/Quick-Start-(Demo)*.

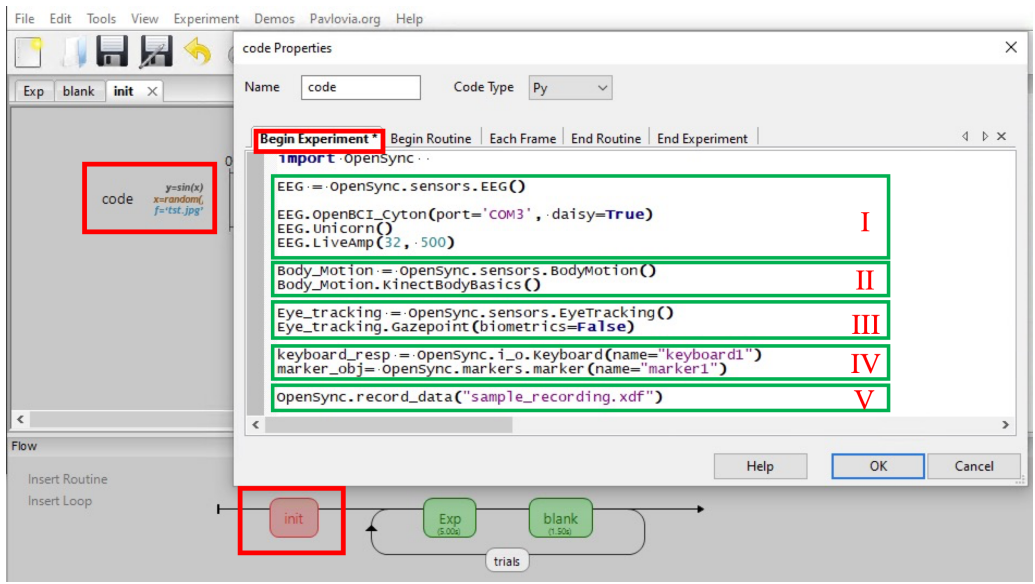

Figure 2.a. Code snippet for initialization of the OpenSync Sensor and Marker modules. Setting up (I) EEG, (II) Kinect body motion, (III) Gazepoint eye-tracking sensors can be done with a few lines. (IV) initialization of I/O and stimulus markers use I/O and Marker modules. (V) with OpenSync.record_data("sample_recording.xdf"), OpenSync starts recording time-stamped data streams and store the data as "sample_recording.xdf".

In Figure 2.a., after importing OpenSync, different devices (EEG, eye-tracking, GSR, body motion and keyboard) and a marker object are initialized and start streaming data. Then, the recording function is called to start recording the available streams. Note that this code snippet is added to PsychoPy's Begin Experiment segment.

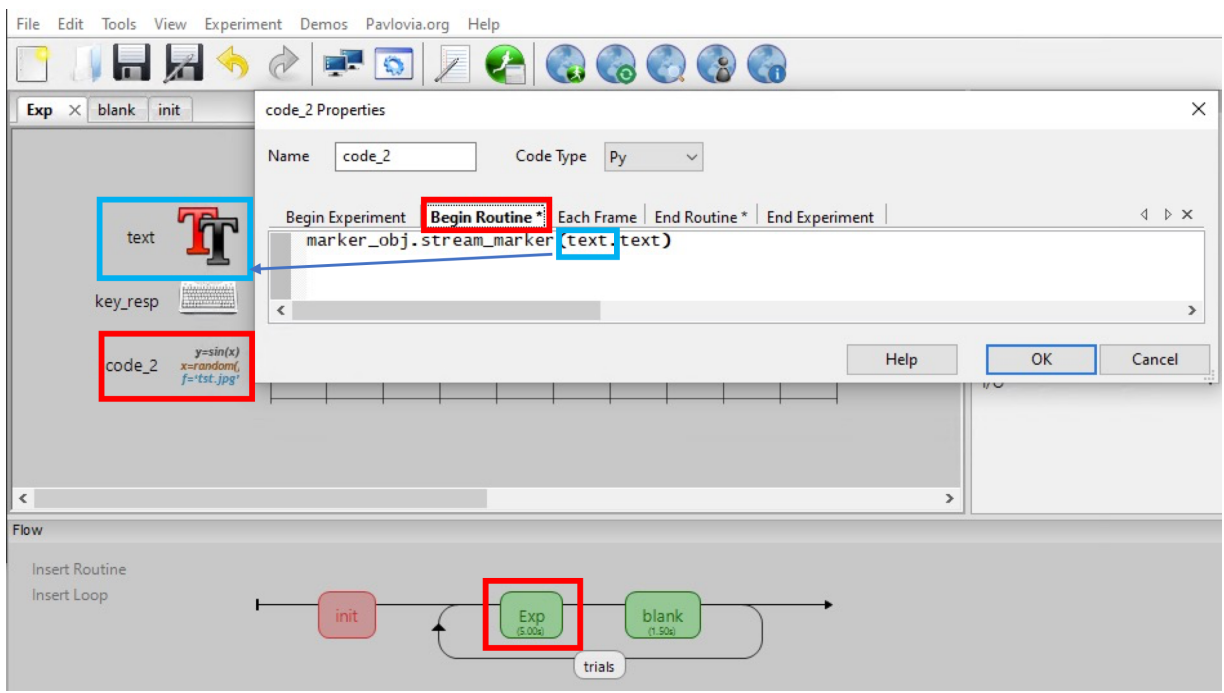

Figure 2.b. Code snippet for streaming stimulus markers

In Figure 2.b., the marker object defined in Figure 2.a. is used to stream the stimulus presentation marker, which is defined by a text object to be shown on the screen.

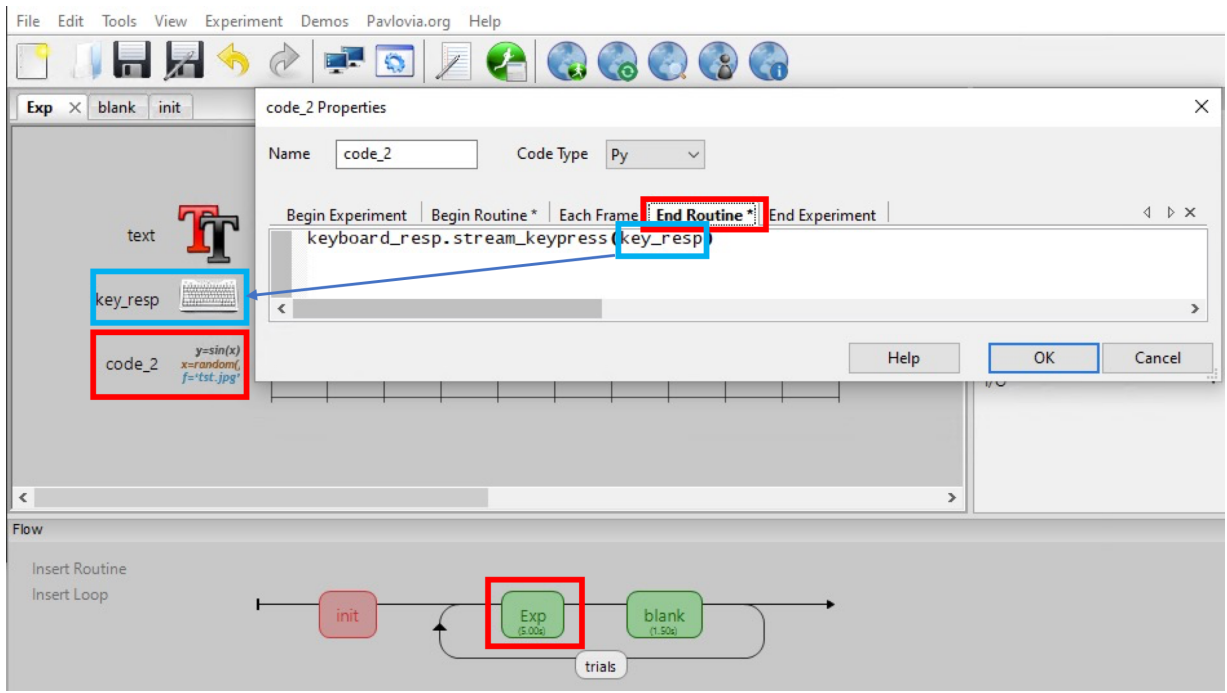

Figure 2.c. Code snippet for streaming user response (I/O) markers

In Figure 2.c., the keyboard object defined in Figure 2.a. is used to stream the response marker associated with the user's key-press event by a keyboard.

As shown in Figure 2, data streams from multiple sources (e.g., EEG, eye-tracking, body motion, stimulus markers, and keyboard response markers) can be synchronized via OpenSync functions embedded in the PsychoPy code interface. All the streams will be recorded in a single *.xdf* file on the disk by the function *OpenSync.record_data* ("file_name.xdf"). Our Github wiki (*https://github.com/TAMUCogLab/OpenSync/wiki/Analyzing-data-(.xdf)*) describes in detail how to analyze .xdf files.

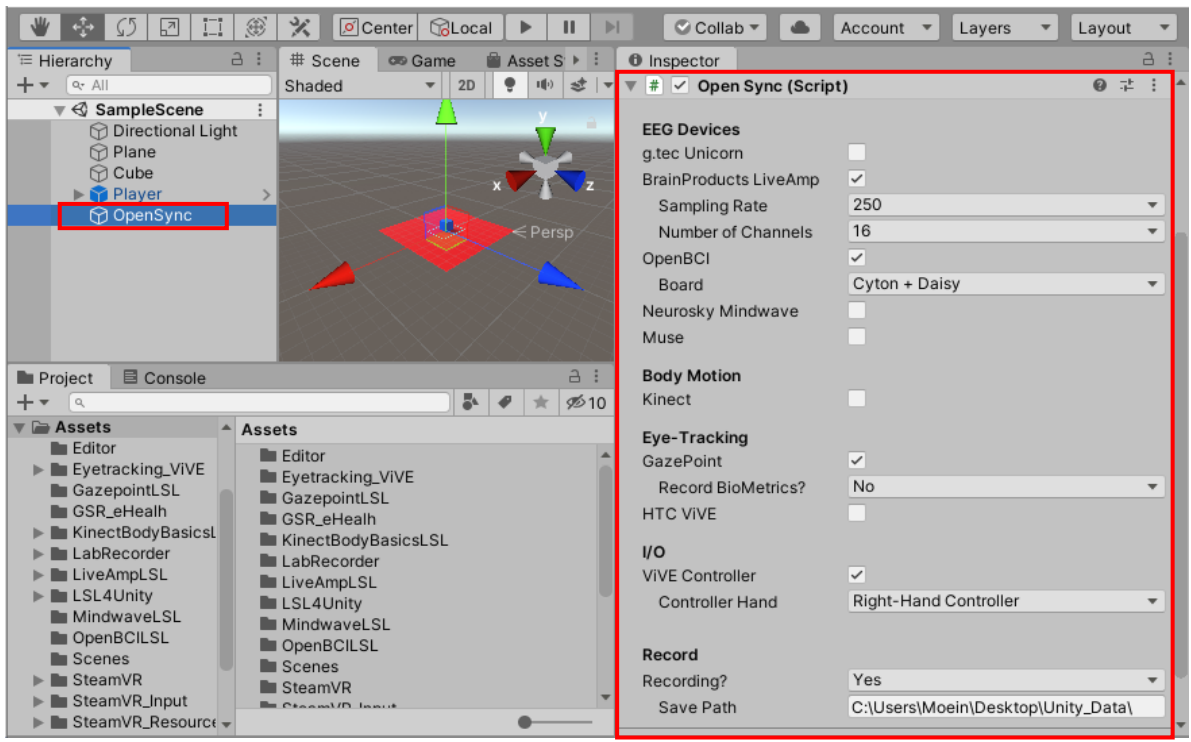

Figure 3. OpenSync in Unity. Once the OpenSync.cs file is attached to the a GameObject (here "OpenSync" in the Hierarchy panel), the OpenSync window would be accessible by the experimenter to select different devices for synchronization.

*3.2. Study 2: Using OpenSync in Unity*

We have designed a Unity inspector that allows the user to stream data from multiple devices and synchronize them through the GUI without writing code. The user can also set device configurations (e.g., in the EEG device BrainProducts LiveAmp, they can set the Sampling Rate and Number of Channels). In order to record the streams, the user only needs to set the output path, and data recording will be done automatically by playing the Scene (Figure 3).

In order to use OpenSync Platform in Unity, all the files and folders available on *https://github.com/TAMUCogLab/OpenSync/tree/master/Demo/Unity* should be added to /Assets/ folder of the project (note that OpenSync.cs and OpenSync_Editor.cs should be added to /Assets/ and /Assets/Editor/ folders, respectively). Then, OpenSync.cs should be attached to a GameObject in the Unity scene. Figure 3. shows OpenSync attached to the GameObject named "OpenSync"; it contains some sample devices and measures. Other devices and measures can be easily added by following the same procedure in the OpenSync scripts available on *https://github.com/TAMUCogLab/OpenSync/tree/master/Demo/Unity/OpenSync.cs* and *https://github.com/TAMUCogLab/OpenSync/tree/master/Demo/Unity/Editor/OpenSync_Editor.cs*.

## 4. Time Synchronization Test

We tested OpenSync's ability for time-synchronization. To that end, we designed an experiment to compare the difference between nominal and extended effective sampling times for a particular device (device working alone versus working with other devices). Specifically, we used g.tec Unicorn as the reference device and then incrementally added OpenBCI, Gazepoint, Kinect, Mouse, and Keyboard devices. In doing so, we examined the extent to which OpenSync's time-synchronization would remain stable over different computational demands in short (5 minutes) and long (60 minutes) periods. This manipulation was intended to test if OpenSync's temporal stability remain robust when delays in data acquisition were accumulated with time.

*4.1 Experiment*

Here we explain in detail the steps of the experiment that we conducted to measure the OpenSync latency resolution. In the experiment, the device sends data to be saved on the disk, and the average sample recording time is measured. Figure 4 shows the flow of this experiment.

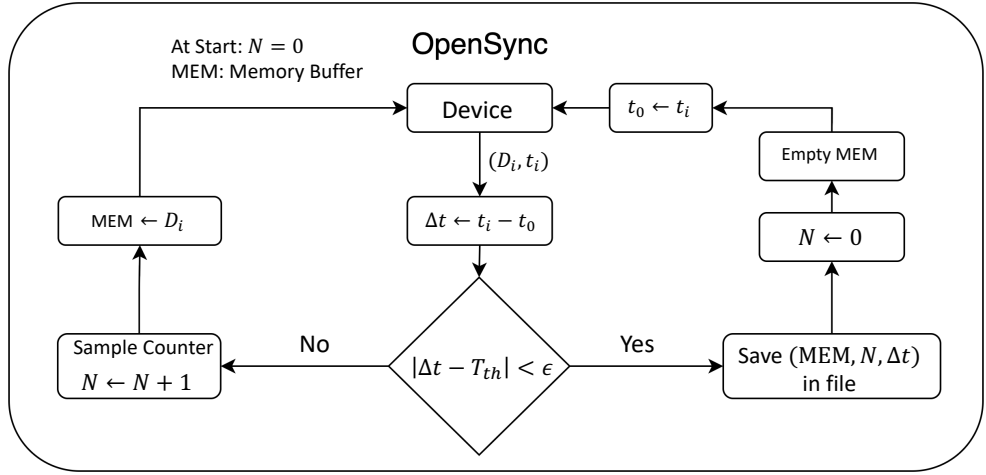

Figure 4. Experiment flowchart. The device starts sending data to the memory buffer. While $|\Delta t - T_{th}| \geq \epsilon$, which means that the time elapsed in filling the current data chunk in the memory buffer has not yet reached $T_{th}$ (for OpenSync $T_{th} = 500ms$), the sample counter counts the number of samples added to the memory buffer ($\Delta t$ is the time elapsed when a chunk of data is being filled). Once $|\Delta t - T_{th}| < \epsilon$, the data samples in the memory buffer (MEM), along with current filled size ($N$) and the time elapsed for that chunk to be filled ($\Delta t$ which is approximately equal to $T_{th}$ when $|\Delta t - T_{th}| < \epsilon$) are saved in a file loaded from disk and the sample counter resets. Then, this process will repeat for the next data chunk.

In this figure:
- $D_i$: $i^{th}$ data sample from the device
- $t_i$: timestamp at which the $i^{th}$ sample enters the memory buffer
- $t_o$: timestamp at which the first sample of each data chunk enters the memory buffer
- $T_{th}$: threshold time set by OpenSync used to determine the time period of each data chunk (In OpenSync, $T_{th}$ is set on 500 ms for each chunk)

The difference between the timestamps of two consecutive loops ($t_{i+1} - t_i$) equals the device's real sampling time (real sampling time contains nominal sampling time plus the delay caused by the device) plus the time lag induced by OpenSync for recording each sample. The following formula derives the average time lag *s* per sample:

$$s = \left(\frac{\Delta t}{N}\right)_{|\Delta t - T_{th}| < \epsilon} - \frac{1}{SRate_{nom}}$$

Where $\left(\frac{\Delta t}{N}\right)_{|\Delta t - T_{th}| < \epsilon}$ equals the average time per sample in the data chunk to be recorded on the disk when the condition $|\Delta t - T_{th}| < \epsilon$ is met. $|\Delta t - T_{th}| < \epsilon$ denotes that the time elapsed in filling a data chunk is very close to $T_{th}$ ($\Delta t$ differs from $T_{th}$ by a very small value of $\epsilon$, in this experiment $\epsilon = 10^{-6}$ secs). Also, $SRate_{nom}$ is the nominal sampling rate of the device. For example, Unicorn's nominal sampling rate is 250 Hz. When $T_{th} = 500\ ms = 0.5\ secs$, then at the end of each data chunk, $\Delta t \approx 0.5\ secs$. As an example, to calculate *s* (average time lag per sample in the chunk), if there are 120 samples ($N$) in the chunk, we have $s = \left(\frac{0.5}{120} - \frac{1}{250}\right) \approx 167 * 10^{-6}\ secs = 167\ \mu s$

According to the flowchart, in the experiment, while the condition $|\Delta t - T_{th}| < \epsilon$ is not met, a counter is enabled, counts the number of data samples from the device and sends the data into the memory buffer. Once the mentioned condition is met, a file would be loaded from the disk and the data samples available in the buffer, the number of data samples, the time elapsed to receive that chunk of samples would be saved in that file and the buffer would be emptied. At this time, the sample counter would reset and the same process repeats.

Four different cases were included in the experiment:
- Case 1 – Unicorn only,
- Case 2 – Unicorn + OpenBCI,
- Case 3 – Unicorn + OpenBCI + Gazepoint,
- Case 4 – Unicorn + OpenBCI + Gazepoint + Kinect + Mouse + Keyboard

*4.2. Results*

In this section, we analyzed the delay caused by the OpenSync platform for synchronization in a Flanker task designed by PsychoPy using a computer system with the following configuration:
- OS: Windows 10 Education
- Processor: Inter® Core™ i7-3612QM CPU @ 2.10 GHz,
- memory (RAM): 12.0 GB (11.9 GB usable),
- System type: 64-bit Operating System, x64 based processor.

Figure 4 shows the distribution (histogram) of the differences between Unicorn nominal sampling time (250 Hz ≡ 4 ms) and the average real recording time per sample (which includes device nominal sample time plus delay induced by the device and the delay caused by OpenSync) in a Flanker task designed in PsychoPy for the four different cases.

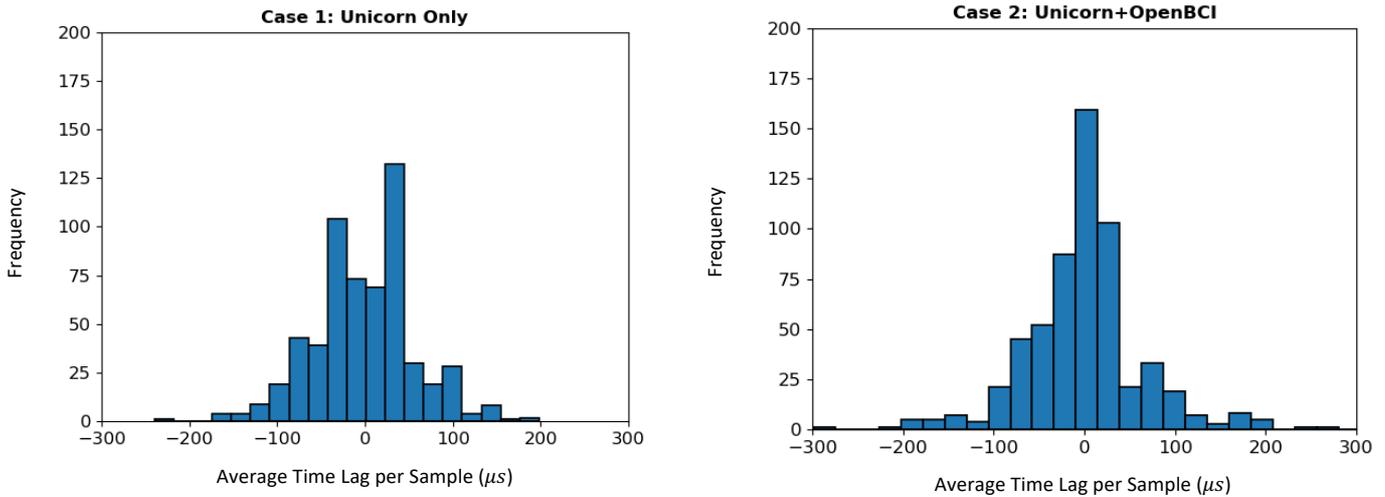

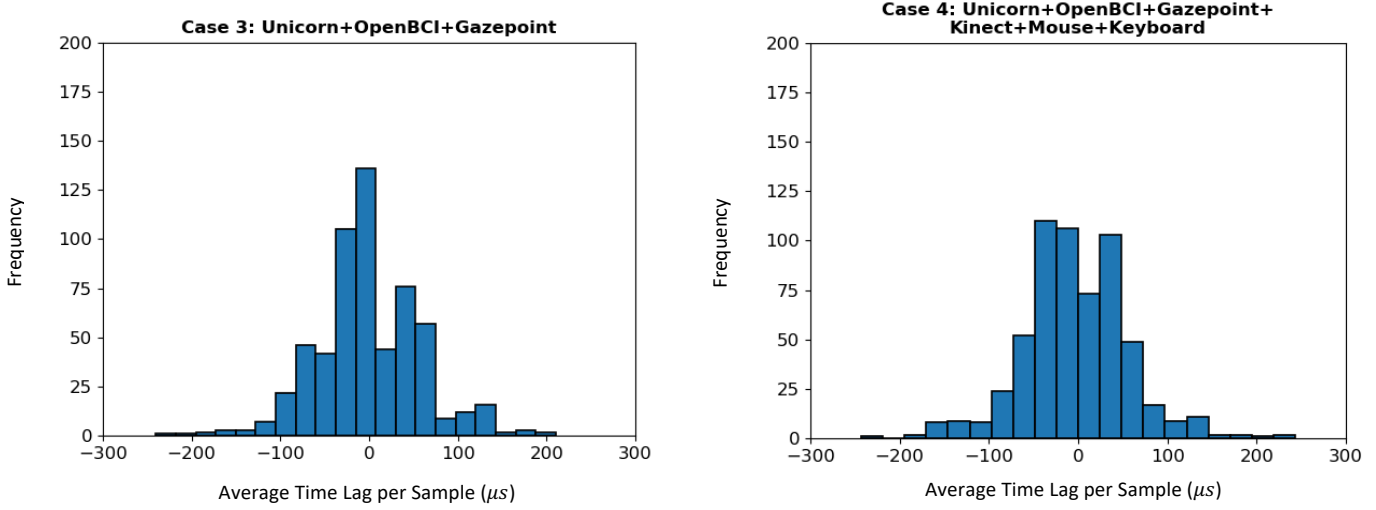

Figure 4. Histograms of average time lag ($s$) for Unicorn device using OpenSync in four different cases with different number of devices

Table 5 shows the descriptive statistics of the experiment.

Table 5. Descriptive statistics of the experiment

|  | Case 1 | | Case 2 | | Case 3 | | Case 4 | |
|---|---|---|---|---|---|---|---|---|
|  | 5 mins | 60 mins | 5 mins | 60 mins | 5 mins | 60 mins | 5 mins | 60 mins |
| N_sample | 600 | 7200 | 600 | 7200 | 600 | 7200 | 600 | 7200 |
| Root-Mean-Square (µs) | 50.69 | 58.47 | 65.28 | 60.29 | 59.39 | 62.33 | 60.16 | 61.72 |
| SD | 57.7 | 58.44 | 65.3 | 65.9 | 59.4 | 62.71 | 60.2 | 61.94 |
| SE | 2.38 | 1.19 | 2.69 | 1.34 | 2.45 | 1.28 | 2.48 | 1.26 |
| Min (µs) | -239.67 | -255.78 | -298.71 | -318.08 | -240.32 | -266.36 | -243.15 | -264.49 |
| Max (µs) | 198.23 | 218.99 | 352.58 | 369.65 | 210.21 | 257.65 | 243.28 | 258.8 |

In this table, each sample is the average time lag of recording Unicorn data samples within each data chunk. In each experiment case, every data chunk includes 500 $ms$ of data recording in two scenarios of 5- and 60-minutes duration. Therefore, we have $n1$ = 600 and $n2$ = 7200 samples for each condition. Root-mean-square (RMS) is suggested by [33] as a measure for reporting time delays. We use RMS of the latencies to compare the latency of OpenSync with other platforms. As we move forward from Case 1 to Case 4 (adding the number of devices), we see that the root mean square of latency only differs by a small fraction and it is not always increasing. In addition, the minimum and maximum time lags are in the microsecond range, which meets our resolution criteria. For the first scenario, the root-mean-square discrepancy is 65.28 µs, and the maximum magnitude of the discrepancy is 352.58 µs. For the second scenario, the root-mean-square discrepancy is 62.33 µs, and the maximum magnitude of the discrepancy is 369.65 µs. The table also shows the root-mean-squares in both scenarios are close to each other, which shows the system's stability. Moreover, as can be seen in the second scenario results, the root-mean-squares of latency in this scenario are close to each other and the standard error values are lower than the first scenario, which indicates that the system reaches a steady state over time.

To compare the time lag between the four conditions we first use Levene's test to check the homogeneity of the variances between the distributions of these conditions [34]. The result show that their variances are homogeneous, $F(3, 2396) = 0.04$, $p = 0.99$. Then, we apply Fisher's One-way ANOVA [35], [36]. No significant difference is found between the time lags in four conditions of the first scenario, $F(3, 2396) = 0.08$, $p = 0.97$, $\eta = 0.6$, $power = 1.0$. This indicates that increasing the number of devices using OpenSync does not affect the effective sampling rate of the devices.

## 5. Discussion

*5.1. Summary*

The ability to use multiple synchronized measures to distinguish factors affecting behavior and brain functionality is getting more attention. Multiple studies showed that using several measures (multimodal experiments) can improve the accuracy and the confidence for interpretation of the results. Synchronizing multiple measures with different sampling rates has various challenges; only a few studies tried to integrate more than two measures. In this paper, we introduce the OpenSync platform for integrating and synchronizing multiple measures. Once the experiment is created using this platform, it is straightforward and time-saving for the experimenter. Every measure starts being recorded and saved on the disk automatically. We also design an experiment to measure the latency resolution of OpenSync and compare it with other platforms.

Compared to other proprietary systems such as iMotion, BioPac, and Neurobs, OpenSync, which is completely free and open-source, outperforms these systems in its extendibility. While OpenSync supports a variety of major programming languages, Python, C, Java, C#, C++, and Matlab, proprietary systems, iMotions and BioPac, barely support these languages for a custom extension. OpenSync also surpasses free/open-source systems (NEDE and USE) in its ease of implementation and flexibility. OpenSync allows off-the-shelf consumer-grade devices (e.g., Neurosky, Muse, Kinect), as well as research-grade high-capacity devices (BrainProducts LiveAmp, OpenBCI Cyton) to be integrated for multimodal and multisensory experimentation while providing good time resolution capacity.

*5.2. Limitation*

There is a limitation with the experiment designed for measuring latency resolution. It records the latency for chunks of data samples and takes the average latency for each sample instead of the absolute latency for each sample. This limitation is due to LSL's sending and receiving data in chunks instead of sample by sample. Moreover, in the experiment that we designed to test the time synchronization, the latency that is measured does not show the latency caused by OpenSync alone; instead, it contains latency caused by OpenSync plus the device latency. Thus, OpenSync's latency resolution partly depends on the base device (e.g., Unicorn) that we used to measure temporal resolution. It is possible that, with different base devices, OpenSync's temporal resolution may change significantly.

Another limitation of this work is that in the current version, it is only possible to record the measurements into a file on the disk and it is not possible to conduct online processing on the data which can be used for BCI and neurofeedback purposes. Currently, OpenSync does not have a GUI (graphical user interface), which allows the user to choose with ease sensor devices that can be integrated into the system.

*5.3. Future work*

The use of OpenSync makes it feasible to record data from multiple subjects. A LAN connection can be used to connect multiple computers, making all the data streams accessible on all the computers. Once all the streams are initialized on all the computers, the record function will record data from multiple computers/subjects. Using that feature, real-time hyperscanning studies will be another type of experiment possible to conduct with OpenSync in the future [37]. This platform can be easily expanded and used for other purposes, such as Brain-Computer Interface and Neurofeedback experiments [38]. Moreover, this platform can be expanded for multimodal-multisensory experiments that involve different human senses (e.g., tactile, hear, smell, taste, etc.) and experiments that are involved with Augmented/Virtual Reality (AR/VR), e.g. [39]–[41]. Applying OpenSync to different human senses will improve accuracy and will likely advance our knowledge of human behavior [42]. For this purpose, the state-of-the-art deep learning models are powerful tools recently used to combine and analyze data from multiple sources together [43].

*5.4. Concluding remarks*

The features of OpenSync presented in this study, when combined, provide a comprehensive framework for synchronizing neuroscience and psychology experiments. An experimenter may start syncing multiple devices and measurements that are available in the platform library using OpenSync. Using the tutorial at *https://github.com/TAMUCogLab/OpenSync/wiki/Extending-OpenSync-for-New-Devices*, one can extend OpenSync for new devices. Using minimum setup time and code modifications, the experimenter may simply expand OpenSync for various devices and measurements not now accessible in the platform, allowing a broad range of neuroscience/psychology studies with multimodal-multisensory measurements.

**Appendix**

*Lab Streaming Layer*

LSL is an Application Programming Interface (API) that is available on open-source platforms (e.g., Python, C, C++, C#, Java, Octave, etc.). It uses Transmission Control Protocol (TCP) for stream transport and User Datagram Protocol (UDP) for stream discovery and time synchronization. TCP is a connection-oriented protocol that guarantees errorless, reliable, and ordered data streaming, and it works with Internet Protocol (IP) for data streaming [44], [45] whereas UDP is a connectionless protocol [46]. The Software Development Kit (SDK) of almost every consumer and research-grade device that supports open-source platforms (e.g., Python, C, C++, C#, Java, Octave, etc.) can be customized to stream data via LSL.

*Installing OpenSync*

OpenSync can be installed for any Python environment using *"pip install OpenSync"*. To install OpenSync on PsychoPy, set the current directory in the command line to PsychoPy Python directory (e.g., use *"cd C:\Program Files\PsychoPy3"* command) and then use *"python -m pip install OpenSync --user"*.

To use OpenSync on Unity, download the files from the *OpenSync_Unity* GitHub repository (*https://github.com/TAMUCogLab/OpenSync/tree/master/Demo/Unity*) and copy them inside the */Assets/* folder in your Unity project. Then create a new GameObject in Unity and add the *OpenSync.cs* script to that GameObject. This will automatically add OpenSync GUI to Unity inspector.

A complete guide on how to install and use OpenSync, including demos is available on: *https://github.com/TAMUCogLab/OpenSync*

*.xdf File Structure*

In order to open the *.xdf* file in Python, first, it is required to install *pyxdf* in python using pip in the command line: *"pip install pyxdf"*. The *.py* file in the link: pyxdf_example, is an example of opening *.xdf* files in Python. It is recommended to use *Spyder* (*https://docs.spyder-ide.org/installation.html*) as the Python platform to open the *.xdf* files since the Variable Explorer panel in *Spyder* allows for tracking the variables. The fields of a *.xdf* file in Python are shown in shown Figure A1. More information is available on: *https://github.com/TAMUCogLab/OpenSync/wiki*

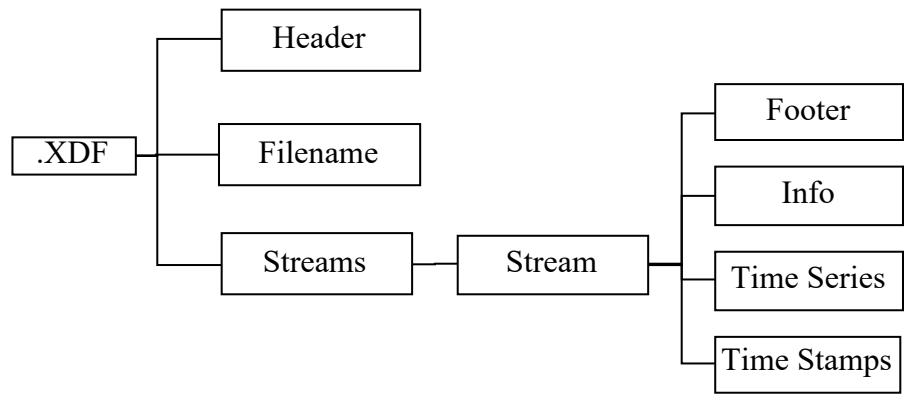

Figure A1. Fields of *.xdf* file

**References**

[1] R. A. Stevenson *et al.*, "Identifying and Quantifying Multisensory Integration: A Tutorial Review," *Brain Topogr.*, vol. 27, no. 6, pp. 707–730, 2014.

[2] T. U. Otto, B. Dassy, and P. Mamassian, "Principles of multisensory behavior," *J. Neurosci.*, vol. 33, no. 17, pp. 7463–7474, 2013.

[3] R. Höchenberger, N. A. Busch, and K. Ohla, "Nonlinear response speedup in bimodal visual-olfactory object identification," *Front. Psychol.*, vol. 6, no. September, pp. 1–11, 2015.

[4] H. D. Critchley *et al.*, "Human cingulate cortex and autonomic control: Converging neuroimaging and clinical evidence," *Brain*, vol. 126, no. 10, pp. 2139–2152, 2003.

[5] J. Kittler, M. Hatef, R. P. W. Duin, and J. Matas, "On combining classifiers," *IEEE Trans. Pattern Anal. Mach. Intell.*, vol. 20, no. 3, pp. 226–239, 1998.

[6] L. M. Reeves, D. D. Schmorrow, and K. M. Stanney, "Augmented cognition and cognitive state assessment technology - Near-term, mid-term, and long-term research objectives," *Lect. Notes Comput. Sci. (including Subser. Lect. Notes Artif. Intell. Lect. Notes Bioinformatics)*, vol. 4565 LNAI, pp. 220–228, 2007.

[7] A. Jimenez-Molina, C. Retamal, and H. Lira, "Using psychophysiological sensors to assess mental workload during web browsing," *Sensors (Switzerland)*, vol. 18, no. 2, pp. 1–26, 2018.

[8] J. Born, B. R. N. Ramachandran, S. A. Romero Pinto, S. Winkler, and R. Ratnam, "Multimodal study of the effects of varying task load utilizing EEG, GSR and eye-tracking," *bioRxiv*, no. 1998, p. 798496, 2019.

[9] Z. Wang, G. Healy, A. F. Smeaton, and T. E. Ward, "An investigation of triggering approaches for the rapid serial visual presentation paradigm in brain computer interfacing," *2016 27th Irish Signals Syst. Conf. ISSC 2016*, pp. 1–6, 2016.

[10] A. Leontyev, T. Yamauchi, and M. Razavi, "Machine Learning Stop Signal Test (ML-SST): ML-based Mouse Tracking Enhances Adult ADHD Diagnosis," in *2019 8th International Conference on Affective Computing and Intelligent Interaction Workshops and Demos, ACIIW 2019*, 2019, pp. 248–252.

[11] A. Leontyev, S. Sun, M. Wolfe, and T. Yamauchi, "Augmented Go/No-Go Task: Mouse Cursor Motion Measures Improve ADHD Symptom Assessment in Healthy College Students," *Front. Psychol.*, vol. 9, no. APR, p. 496, Apr. 2018.

[12] A. Leontyev and T. Yamauchi, "Mouse movement measures enhance the stop-signal task in adult ADHD assessment," *PLoS One*, vol. 14, no. 11, p. e0225437, 2019.

[13] T. Yamauchi and K. Xiao, "Reading Emotion From Mouse Cursor Motions: Affective Computing Approach," *Cogn. Sci.*, vol. 42, no. 3, pp. 771–819, Apr. 2018.

[14] T. Yamauchi, A. Leontyev, and M. Razavi, "Assessing Emotion by Mouse-cursor Tracking: Theoretical and Empirical Rationales," in *2019 8th International Conference on Affective Computing and Intelligent Interaction, ACII 2019*, 2019, pp. 89–95.

[15] T. Yamauchi, J. Seo, and A. Sungkajun, "Interactive Plants: Multisensory Visual-Tactile Interaction Enhances Emotional Experience," *Mathematics*, vol. 6, no. 11, p. 225, Oct. 2018.

[16] M. Razavi, T. Yamauchi, V. Janfaza, A. Leontyev, S. Longmire-Monford, and J. Orr, "Multimodal-Multisensory Experiments." Preprints, 12-Oct-2020.

[17] F. Chen *et al.*, "Multimodal behavior and interaction as indicators of cognitive load," *ACM Trans. Interact. Intell. Syst.*, vol. 2, no. 4, 2012.

[18] N. Lazzeri, D. Mazzei, and D. De Rossi, "Development and Testing of a Multimodal Acquisition

Platform for Human-Robot Interaction Affective Studies," *J. Human-Robot Interact.*, vol. 3, no. 2, p. 1, 2014.

[19] P. Cornelio, C. Velasco, and M. Obrist, "Multisensory Integration as per Technological Advances: A Review," *Front. Neurosci.*, vol. 15, Jun. 2021.

[20] R. L. Charles and J. Nixon, "Measuring mental workload using physiological measures: A systematic review," *Applied Ergonomics*, vol. 74. Elsevier Ltd, pp. 221–232, 01-Jan-2019.

[21] M. Lohani, B. R. Payne, and D. L. Strayer, "A review of psychophysiological measures to assess cognitive states in real-world driving," *Front. Hum. Neurosci.*, vol. 13, no. March, pp. 1–27, 2019.

[22] E. Levin, W. Roland, R. Habibi, Z. An, and R. Shults, "RAPID VISUAL PRESENTATION TO SUPPORT GEOSPATIAL BIG DATA PROCESSING," 2020.

[23] B. E. Gibson *et al.*, "A multi-method approach to studying activity setting participation: Integrating standardized questionnaires, qualitative methods and physiological measures," *Disabil. Rehabil.*, vol. 36, no. 19, pp. 1652–1660, 2014.

[24] L. W. Sciarini and D. Nicholson, "Assessing cognitive state with multiple physiological measures: A modular approach," *Lect. Notes Comput. Sci. (including Subser. Lect. Notes Artif. Intell. Lect. Notes Bioinformatics)*, vol. 5638 LNAI, pp. 533–542, 2009.

[25] "iMotions." [Online]. Available: https://imotions.com/. [Accessed: 29-May-2021].

[26] "Data Acquisition, Loggers, Amplifiers, Transducers, Electrodes | BIOPAC." [Online]. Available: https://www.biopac.com/. [Accessed: 29-May-2021].

[27] "Neurobehavioral Systems." [Online]. Available: https://www.neurobs.com/.

[28] D. C. Jangraw, A. Johri, M. Gribetz, and P. Sajda, "NEDE: An open-source scripting suite for developing experiments in 3D virtual environments," *J. Neurosci. Methods*, vol. 235, pp. 245–251, 2014.

[29] M. R. Watson, B. Voloh, C. Thomas, A. Hasan, and T. Womelsdorf, "USE: An integrative suite for temporally-precise psychophysical experiments in virtual environments for human, nonhuman, and artificially intelligent agents," *J. Neurosci. Methods*, vol. 326, no. June, 2019.

[30] "AcqKnowledge Software Guide For Life Science Research Applications." [Online]. Available: https://www.biopac.com/wp-content/uploads/acqknowledge_software_guide.pdf.

[31] "neurobs." [Online]. Available: https-//www.neurobs.com/menu_presentation/menu_hardware/labstreamer.

[32] "Introduction — Labstreaminglayer 1.13 documentation." [Online]. Available: https://labstreaminglayer.readthedocs.io/info/intro.html. [Accessed: 31-May-2021].

[33] O. Ur-Rehman and N. Zivic, *Wireless communications*. 2018.

[34] H. Levene and others, "Contributions to probability and statistics," *Essays Honor Harold Hotell.*, pp. 278–292, 1960.

[35] R. A. Fisher, "XV.—The Correlation between Relatives on the Supposition of Mendelian Inheritance," *Trans. R. Soc. Edinburgh*, vol. 52, no. 2, pp. 399–433, 1919.

[36] R. A. Fisher, "On the 'probable error' of a coefficient of correlation deduced from a small sample," *Metron*, vol. 1, pp. 1–32, 1921.

[37] A. Czeszumski *et al.*, "Hyperscanning: A Valid Method to Study Neural Inter-brain Underpinnings of Social Interaction," *Front. Hum. Neurosci.*, vol. 0, p. 39, Feb. 2020.

[38] R. Abiri, S. Borhani, E. W. Sellers, Y. Jiang, and X. Zhao, "A comprehensive review of EEG-based brain-computer interface paradigms," *Journal of Neural Engineering*, vol. 16, no. 1. Institute of Physics Publishing, p. 011001, 01-Feb-2019.

[39] E. Levin, R. Shults, R. Habibi, Z. An, and W. Roland, "Geospatial Virtual Reality for Cyberlearning in the Field of Topographic Surveying: Moving Towards a Cost-Effective Mobile Solution," *ISPRS Int. J. Geo-Information 2020, Vol. 9, Page 433*, vol. 9, no. 7, p. 433, Jul. 2020.

[40] S. R. K. Tabbakh, R. Habibi, and S. Vafadar, "Design and implementation of a framework based on augmented reality for phobia treatment applications," *2nd Int. Congr. Technol. Commun. Knowledge, ICTCK 2015*, pp. 366–370, Oct. 2016.

[41] R. Habibi, "Detecting Surface Interactions via a Wearable Microphone to Improve Augmented Reality Text Entry," *Diss. Master's Theses Master's Reports*, Jan. 2021.

[42] E. Marsja, J. E. Marsh, P. Hansson, and G. Neely, "Examining the role of spatial changes in bimodal and uni-modal to-be-ignored stimuli and how they affect short-term memory processes," *Front. Psychol.*, vol. 10, no. MAR, pp. 1–8, 2019.

[43] J. Gao, P. Li, Z. Chen, and J. Zhang, "A survey on deep learning for multimodal data fusion," *Neural Comput.*, vol. 32, no. 5, pp. 829–864, 2020.

[44] J. Postel, "DOD standard transmission control protocol," *ACM SIGCOMM Comput. Commun. Rev.*, vol. 10, no. 4, pp. 52–132, 1980.

[45]  W. R. Stevens, *TCP/IP Illustrated (Vol. 1): The Protocols*. USA: Addison-Wesley Longman Publishing Co., Inc., 1993.
[46]  P. Loshin, "User Datagram Protocol," *TCP/IP Clear. Explain.*, no. August, pp. 341–349, 2003.